\documentclass[10pt,journal,epsfig]{IEEEtran}

\usepackage[dvips]{graphicx}
\usepackage{graphicx}
\usepackage{amssymb}
\usepackage{cite}
\usepackage{subfigure}
\usepackage{amsmath}

\begin{document}


\title{Robust Bayesian Compressed Sensing}

\author{Qian Wan, Huiping Duan, Jun Fang, and Hongbin Li,~\IEEEmembership{Senior Member,~IEEE}
\thanks{Qian Wan and Jun Fang are with the National Key Laboratory
of Science and Technology on Communications, University of
Electronic Science and Technology of China, Chengdu 611731, China,
Email: JunFang@uestc.edu.cn}
\thanks{Huiping Duan is with the School of Electronic Engineering,
University of Electronic Science and Technology of China, Chengdu
611731, China, Email: huipingduan@uestc.edu.cn}
\thanks{Hongbin Li is
with the Department of Electrical and Computer Engineering,
Stevens Institute of Technology, Hoboken, NJ 07030, USA, E-mail:
Hongbin.Li@stevens.edu}
\thanks{This work was supported in part by the National Science
Foundation of China under Grant 61522104, and the National Science
Foundation under Grant ECCS-1408182 and Grant ECCS-1609393, and
the Air Force Office of Scientific Research under Grant
FA9550-16-1-0243.}}

\maketitle



\begin{abstract}
We consider the problem of robust compressed sensing where the
objective is to recover a high-dimensional sparse signal from
compressed measurements partially corrupted by outliers. A new
sparse Bayesian learning method is developed for this purpose. The
basic idea of the proposed method is to identify the outliers and
exclude them from sparse signal recovery. To automatically
identify the outliers, we employ a set of binary indicator
variables to indicate which observations are outliers. These
indicator variables are assigned a beta-Bernoulli hierarchical
prior such that their values are confined to be binary. In
addition, a Gaussian-inverse Gamma prior is imposed on the sparse
signal to promote sparsity. Based on this hierarchical prior
model, we develop a variational Bayesian method to estimate the
indicator variables as well as the sparse signal. Simulation
results show that the proposed method achieves a substantial
performance improvement over existing robust compressed sensing
techniques.
\end{abstract}



\section{Introduction}
Compressed sensing, a new paradigm for data acquisition and
reconstruction, has drawn much attention over the past few years
\cite{ChenDonoho98,CandesTao05,Donoho06}. The main purpose of
compressed sensing is to recover a high-dimensional sparse signal
from a low-dimensional linear measurement vector. In practice,
measurements are inevitably contaminated by noise due to hardware
imperfections, quantization errors, or transmission errors. Most
existing studies (e.g.
\cite{Candes08,Wainwright09,WimalajeewaVarshney12}) assume that
measurements are corrupted with noise that is evenly distributed
across the observations, such as independent and identically
distributed (i.i.d.) Gaussian, thermal, or quantization noise.
This assumption is valid for many cases. Nevertheless, for some
scenarios, measurements may be corrupted by outliers that are
significantly different from their nominal values. For example,
during the data acquisition process, outliers can be caused by
sensor failures or calibration errors
\cite{LaskaDavenport09,MitraVeeraraghavan13}, and it is usually
unknown which measurements have been corrupted. Outliers can also
arise as a result of signal clipping/saturation or impulse noise
\cite{CarrilloBarner10,StuderKuppinger12}. Conventional compressed
sensing techniques may incur severe performance degradation in the
presence of outliers. To address this issue, in previous works
(e.g.
\cite{LaskaDavenport09,MitraVeeraraghavan13,CarrilloBarner10,StuderKuppinger12}),
outliers are modeled as a sparse error vector, and the observed
data are expressed as
\begin{align}
\boldsymbol{y}=\boldsymbol{A}\boldsymbol{x}+\boldsymbol{e}+\boldsymbol{w}
\label{RCS}
\end{align}
where $\boldsymbol{A}\in\mathbb{R}^{M\times N}$ is the sampling
matrix with $M\ll N$, $\boldsymbol{x}$ denotes an $N$-dimensional
sparse vector with only $K$ nonzero coefficients,
$\boldsymbol{e}\in\mathbb{R}^{M}$ denotes the outlier vector
consisting of $T\ll M$ nonzero entries with arbitrary amplitudes,
and $\boldsymbol{w}$ denotes the additive multivariate Gaussian
noise with zero mean and covariance matrix
$(1/\gamma)\boldsymbol{I}$. The above model can be formulated as a
conventional compressed sensing problem as
\begin{align}
\boldsymbol{y}= \left[
\begin{array}{cc}\boldsymbol{A}&\boldsymbol{I}\end{array} \right]
\left[
\begin{array}{c}\boldsymbol{x}\\
\boldsymbol{e}\end{array} \right]+\boldsymbol{w}
\triangleq\boldsymbol{B}\boldsymbol{u} +\boldsymbol{w}
\label{RCS-compensation}
\end{align}
Efficient compressed sensing algorithms can then be employed to
estimate the sparse signal as well as the outliers. Recovery
guarantees of $\boldsymbol{x}$ and $\boldsymbol{e}$ were also
analyzed in
\cite{LaskaDavenport09,MitraVeeraraghavan13,CarrilloBarner10,StuderKuppinger12}.



The rationale behind the above approach is to detect and
compensate for these outliers simultaneously. Besides the above
method, another more direct approach is to identify and exclude
the outliers from sparse signal recovery. Although it may seem
preferable to compensate rather than simply reject outliers,
inaccurate estimation of the compensation (i.e. outlier vector)
could result in a destructive effect on sparse signal recovery,
particularly when the number of measurements is limited. In this
case, identifying and rejecting outliers could be a more sensible
strategy. Motivated by this insight, we develop a Bayesian
framework for robust compressed sensing, in which a set of binary
indicator variables are employed to indicate which observations
are outliers. These variables are assigned a beta-Bernoulli
hierarchical prior such that their values are confined to be
binary. Also, a Gaussian inverse-Gamma prior is placed on the
sparse signal to promote sparsity. A variational Bayesian method
is developed to find the approximate posterior distributions of
the indicators, the sparse signal and other latent variables.
Simulation results show that the proposed method achieves a
substantial performance improvement over the compensation-based
robust compressed sensing method.

\begin{figure}[!t]
\centering
\includegraphics[width=5cm]{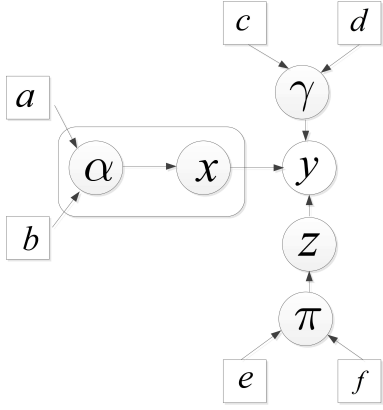}
\caption{Graphical model for robust Bayesian compressed sensing.}
\label{fig3}
\end{figure}

\section{Hierarchical Prior Model}
We develop a Bayesian framework which employs a set of indicator
variables $\boldsymbol{z}\triangleq\{z_m\}$ to indicate which
observation is an outlier, i.e. $z_m=1$ indicates that $y_m$ is a
normal observation, otherwise $y_m$ is an outlier. More precisely,
we can write
\begin{align}
y_m=\begin{cases} \boldsymbol{a}_{m}^r\boldsymbol{x}+w_m & z_m=1
\\ \boldsymbol{a}_{m}^r\boldsymbol{x}+w_m+e_m & z_m=0 \end{cases}
\end{align}
where $\boldsymbol{a}_{m}^r$ denotes the $m$th row of
$\boldsymbol{A}$, $e_m$ and $w_m$ are the $m$th entry of
$\boldsymbol{e}$ and $\boldsymbol{w}$, respectively. The
probability of the observed data conditional on these indicator
variables can be expressed as
\begin{align}
p(\boldsymbol{y}|\boldsymbol{x},\boldsymbol{z},\gamma)=\prod_{m=1}^M
(\mathcal{N}(y_m|\boldsymbol{a}_{m}^r\boldsymbol{x},1/\gamma))^{z_m}
\end{align}
in which those ``presumed outliers'' are automatically disabled
when calculating the probability. To infer the indicator
variables, a beta-Bernoulli hierarchical prior
\cite{HeCarin09,PaisleyCarin09} is placed on $\boldsymbol{z}$,
i.e. each component of $\boldsymbol{z}$ is assumed to be drawn
from a Bernoulli distribution parameterized by $\pi_m$
\begin{align}
p(z_m|\pi_m)=\text{Bernoulli}(z_m|\pi_m)=\pi_m^{z_m}(1-\pi_m)^{1-z_m}
\quad \forall m
\end{align}
and $\pi_m$ follows a beta distribution
\begin{align}
p(\pi_m)=\text{Beta}(e,f) \quad \forall m
\end{align}
where $e$ and $f$ are parameters characterizing the beta
distribution. Note that the beta-Bernoulli prior assumes the
random variables $\{z_m\}$ are mutually independent, and so are
the random variables $\{\pi_m\}$.


To encourage a sparse solution, a Gaussian-inverse Gamma
hierarchical prior, which has been widely used in sparse Bayesian
learning (e.g. \cite{JiXue08,ZhangRao13,YangXie13,FangShen15}), is
employed. Specifically, in the first layer, $\boldsymbol{x}$ is
assigned a Gaussian prior distribution
\begin{align}
p(\boldsymbol{x}|\boldsymbol{\alpha})=\prod_{n=1}^N
p(x_n|\alpha_n)
\end{align}
where $p(x_n|\alpha_n)=\mathcal{N}(x_n|0,\alpha_n^{-1})$, and
$\boldsymbol{\alpha}\triangleq\{\alpha_n\}$ are non-negative
hyperparameters controlling the sparsity of the signal
$\boldsymbol{x}$. The second layer specifies Gamma distributions
as hyperpriors over the precision parameters $\{\alpha_n\}$, i.e.
\begin{align}
p(\boldsymbol{\alpha})=\prod_{n=1}^N\text{Gamma}(\alpha_n|a,b)=\prod_{n=1}^N
\Gamma(a)^{-1}b^a\alpha_n^{a-1}e^{-b\alpha_n} \label{alpha-prior}
\end{align}
where the parameters $a$ and $b$ are set to small values (e.g.
$a=b=10^{-10}$) in order to provide non-informative (over a
logarithmic scale) hyperpriors over $\{\alpha_n\}$. Also, to
estimate the noise variance, we place a Gamma hyperprior over
$\gamma$, i.e.
\begin{align}
p(\gamma)=\text{Gamma}(\gamma|c,d)=\Gamma(c)^{-1}d^c\gamma^{c-1}e^{-d\gamma}
\label{gamma-prior}
\end{align}
where the parameters $c$ and $d$ are set to be small, e.g.
$c=d=10^{-10}$. The graphical model of the proposed hierarchical
prior is shown in Fig. \ref{fig3}.


\section{Variational Bayesian Inference}
We now proceed to perform Bayesian inference for the proposed
hierarchical model. Let
$\boldsymbol{\theta}\triangleq\{\boldsymbol{z},\boldsymbol{x},\boldsymbol{\pi},\boldsymbol{\alpha},\gamma\}$
denote the hidden variables in our hierarchical model. Our
objective is to find the posterior distribution
$p(\boldsymbol{\theta}|\boldsymbol{y})$, which is usually
computationally intractable. To circumvent this difficulty,
observe that the marginal probability of the observed data can be
decomposed into two terms
\begin{align}
\ln p(\boldsymbol{y})=L(q)+\text{KL}(q|| p)
\label{variational-decomposition}
\end{align}
where
\begin{align}
L(q)=\int q(\boldsymbol{\theta})\ln
\frac{p(\boldsymbol{y},\boldsymbol{\theta})}{q(\boldsymbol{\theta})}d\boldsymbol{\theta}
\end{align}
and
\begin{align}
\text{KL}(q|| p)=-\int q(\boldsymbol{\theta})\ln
\frac{p(\boldsymbol{\theta}|\boldsymbol{y})}{q(\boldsymbol{\theta})}d\boldsymbol{\theta}
\end{align}
where $q(\boldsymbol{\theta})$ is any probability density
function, $\text{KL}(q|| p)$ is the Kullback-Leibler divergence
between $p(\boldsymbol{\theta}|\boldsymbol{y})$ and
$q(\boldsymbol{\theta})$. Since $\text{KL}(q|| p)\geq 0$, it
follows that $L(q)$ is a rigorous lower bound on $\ln
p(\boldsymbol{y})$. Moreover, notice that the left hand side of
(\ref{variational-decomposition}) is independent of
$q(\boldsymbol{\theta})$. Therefore maximizing $L(q)$ is
equivalent to minimizing $\text{KL}(q|| p)$, and thus the
posterior distribution $p(\boldsymbol{\theta}|\boldsymbol{y})$ can
be approximated by $q(\boldsymbol{\theta})$ through maximizing
$L(q)$. Specifically, we could assume some specific parameterized
functional form for $q(\boldsymbol{\theta})$ and then maximize
$L(q)$ with respect to the parameters of the distribution. A
particular form of $q(\boldsymbol{\theta})$ that has been widely
used with great success is the factorized form over the component
variables in $\boldsymbol{\theta}$ \cite{TzikasLikas08}. For our
case, the factorized form of $q(\boldsymbol{\theta})$ can be
written as
\begin{align}
q(\boldsymbol{\theta})=q_{z}(\boldsymbol{z})q_{x}(\boldsymbol{x})q_{\alpha}(\boldsymbol{\alpha})q_{\pi}(\boldsymbol{\pi})
q_{\gamma}(\gamma)
\end{align}
We can compute the posterior distribution approximation by finding
$q(\boldsymbol{\theta})$ of the factorized form that maximizes the
lower bound $L(q)$. The maximization can be conducted in an
alternating fashion for each latent variable, which leads to
\cite{TzikasLikas08}
\begin{align}
\ln q_x(\boldsymbol{x})=&\langle\ln
p(\boldsymbol{y},\boldsymbol{\theta})\rangle_{q_{\alpha}(\boldsymbol{\alpha})
q_{\gamma}(\gamma)q_{z}(\boldsymbol{z})q_{\pi}(\boldsymbol{\pi})} + \text{constant} \nonumber\\
\ln q_{\alpha}(\boldsymbol{\alpha})=&\langle\ln
p(\boldsymbol{y},\boldsymbol{\theta})\rangle_{q_x(\boldsymbol{x})
q_{\gamma}(\gamma)q_{z}(\boldsymbol{z})q_{\pi}(\boldsymbol{\pi})} + \text{constant} \nonumber\\
\ln q_{\gamma}(\gamma)=&\langle\ln
p(\boldsymbol{y},\boldsymbol{\theta})\rangle_{q_x(\boldsymbol{x})
q_{\alpha}(\boldsymbol{\alpha})q_{z}(\boldsymbol{z})q_{\pi}(\boldsymbol{\pi})} + \text{constant} \nonumber\\
\ln q_z(\boldsymbol{z})=&\langle\ln
p(\boldsymbol{y},\boldsymbol{\theta})\rangle_{q_x(\boldsymbol{x})
q_{\alpha}(\boldsymbol{\alpha})q_{\gamma}(\gamma)q_{\pi}(\boldsymbol{\pi})} + \text{constant} \nonumber\\
\ln q_{\pi}(\boldsymbol{\pi})=&\langle\ln
p(\boldsymbol{y},\boldsymbol{\theta})\rangle_{q_x(\boldsymbol{x})
q_{\alpha}(\boldsymbol{\alpha})q_{\gamma}(\gamma)q_{z}(\boldsymbol{z})} + \text{constant} \nonumber\\
\label{Q-update}
\end{align}
where $\langle\cdot\rangle_{\cdot}$ denotes an expectation with
respect to the distributions specified in the subscript. More
details of the Bayesian inference are provided below.

\textbf{1) Update of $q_x(\boldsymbol{x})$}: We first consider the
calculation of $q_x(\boldsymbol{x})$. Keeping those terms that are
dependent on $\boldsymbol{x}$, we have
\begin{align}
\ln q_x(\boldsymbol{x})\propto& \langle\ln
p(\boldsymbol{y}|\boldsymbol{x},\boldsymbol{z},\gamma)+\ln
p(\boldsymbol{x}|\boldsymbol{\alpha})\rangle
_{q_{\alpha}(\boldsymbol{\alpha})q_{\gamma}(\gamma)q_z(\boldsymbol{z})} \nonumber\\
\propto& -\sum_{m=1}^M\frac{\langle\gamma
z_m(y_m-\boldsymbol{a}_{m}^r\boldsymbol{x})^{2}\rangle}{2}
-\frac{1}{2}\sum_n^N\langle\alpha_n x_n^{2}\rangle  \nonumber\\
=&
-\frac{\langle\gamma\rangle(\boldsymbol{y}-\boldsymbol{A}\boldsymbol{x})^T
\boldsymbol{D}_{z}(\boldsymbol{y}-\boldsymbol{A}\boldsymbol{x})}{2}
-\frac{1}{2}\boldsymbol{x}^T\boldsymbol{D}_{\alpha}\boldsymbol{x}
\end{align}
where
\begin{align}
\boldsymbol{D}_{z}\triangleq\text{diag}(\langle{\boldsymbol{z}}\rangle),
~~
\boldsymbol{D}_{\alpha}\triangleq\text{diag}(\langle{\boldsymbol{\alpha}}\rangle)
\label{D-update-1}
\end{align}
$\langle{\boldsymbol{z}}\rangle$ and
$\langle{\boldsymbol{\alpha}}\rangle$ denote the expectation of
$\boldsymbol{z}$ and $\boldsymbol{\alpha}$, respectively. It is
easy to show that $q(\boldsymbol{x})$ follows a Gaussian
distribution with its mean and covariance matrix given
respectively by
\begin{align}
\boldsymbol{\mu}_{x}
=&\langle\gamma\rangle\boldsymbol{\Phi}_{x}\boldsymbol{A}^T\boldsymbol{D}_z\boldsymbol{y}
\label{x-update-1}\\
\boldsymbol{\Phi}_{x}
=&\left(\langle\gamma\rangle\boldsymbol{A}^T\boldsymbol{D}_{z}\boldsymbol{A}+
\boldsymbol{D}_{\alpha}\right)^{-1} \label{x-update-2}
\end{align}

\textbf{2) Update of $q_{\alpha}(\boldsymbol{\alpha})$}: Keeping
only the terms that depend on $\boldsymbol{\alpha}$, the
variational optimization of $q_{\alpha}(\boldsymbol{\alpha})$
yields
\begin{align}
\ln q_{\alpha}(\boldsymbol{\alpha}) \propto& \langle\ln
p(\boldsymbol{x}|\boldsymbol{\alpha})+\ln
p(\boldsymbol{\alpha}|a,b)\rangle_{q_x(\boldsymbol{x})} \nonumber\\
=& \sum_{n=1}^N (a+0.5)\ln\alpha_n-(0.5 \left\langle
x_n^2\right\rangle+b)\alpha_n
\end{align}
The posterior $q_{\alpha}(\boldsymbol{\alpha})$ therefore follows
a Gamma distribution
\begin{align}
q_{\alpha}(\boldsymbol{\alpha})=\prod_{n=1}^N\text{Gamma}(\alpha_n|\tilde{a},\tilde{b}_n)
\label{alpha-update}
\end{align}
in which $\tilde{a}$ and $\tilde{b}_n$ are given respectively as
\begin{align}
\tilde{a}=&a+0.5   \nonumber \\
\tilde{b}_n=&b+0.5\langle x_n^{2}\rangle \nonumber
\end{align}


\textbf{3). Update of $q_{\gamma}(\gamma)$:} The variational
approximation of $q_{\gamma}(\gamma)$ can be obtained as:
\begin{align}
\ln q_{\gamma}(\gamma)\propto& \langle\ln
p(\boldsymbol{y}|\boldsymbol{x},\boldsymbol{z},\gamma)+\ln
p(\gamma|c,d)
\rangle_{q_x(\boldsymbol{x})q_z(\boldsymbol{z})} \nonumber\\
\propto& \sum_{m=1}^M \left( 0.5 \langle
z_m\rangle\ln\gamma-0.5\gamma\langle z_m\rangle
\langle(y_m-\boldsymbol{a}_m^r\boldsymbol{x})^2\rangle\right) \nonumber\\
&+(c-1)\ln\gamma-d\gamma  \nonumber\\
=& (c+0.5\sum_{m=1}^M \langle z_m\rangle-1)\ln\gamma-
(d+0.5\langle(\boldsymbol{y}-\boldsymbol{A}\boldsymbol{x})^T  \nonumber\\
&\boldsymbol{D_z}(\boldsymbol{y}-\boldsymbol{A}\boldsymbol{x})\rangle)\gamma
\end{align}
Clearly, the posterior $q_{\gamma}(\gamma)$ obeys a Gamma
distribution
\begin{align}
q_{\gamma}(\gamma)=\text{Gamma}(\gamma|\tilde{c},\tilde{d})
\label{gamma-update}
\end{align}
where $\tilde{c}$ and $\tilde{d}$ are given respectively as
\begin{align}
\tilde{c}=&c+0.5\sum_{m=1}^M \langle z_m\rangle
\\
\tilde{d}=&d+0.5\langle(\boldsymbol{y}-\boldsymbol{A}\boldsymbol{x})^T\boldsymbol{D}_{z}
(\boldsymbol{y}-\boldsymbol{A}\boldsymbol{x})\rangle_{q_{x}(\boldsymbol{x})}
\end{align}
in which
\begin{align}
&\langle(\boldsymbol{y}-\boldsymbol{A}\boldsymbol{x})^T\boldsymbol{D}_{z}
(\boldsymbol{y}-\boldsymbol{A}\boldsymbol{x})\rangle_{q_{x}(\boldsymbol{x})}
\nonumber\\
=&(\boldsymbol{y}-\boldsymbol{A}\boldsymbol{\mu}_{x})^T\boldsymbol{D}_{z}
(\boldsymbol{y}-\boldsymbol{A}\boldsymbol{\mu}_{x})+
\text{trace}(\boldsymbol{A}^T\boldsymbol{D}_{z}\boldsymbol{A}\boldsymbol{\Phi}_{x})
\nonumber
\end{align}


\textbf{4) Update of $q_z(\boldsymbol{z})$:} The posterior
approximation of $q_z(\boldsymbol{z})$ yields
\begin{align}
\ln q_z(\boldsymbol{z})\propto&\langle\ln
p(\boldsymbol{y}|\boldsymbol{x},\boldsymbol{z},\gamma)+\ln
p(\boldsymbol{z}|\boldsymbol{\pi})\rangle_
{q_x(\boldsymbol{x})q_{\gamma}(\gamma)q_{\pi}(\boldsymbol{\pi})}
\nonumber\\
\propto&
 \sum_{m=1}^M\langle z_m\left( -0.5\gamma(y_m-\boldsymbol{a}_m^r\boldsymbol{x})^2+\ln\pi_m\right)+  \nonumber\\
 &(1-z_m)\ln(1-\pi_m)\rangle
\end{align}
Clearly, $z_m$ still follows a Bernoulli distribution with its
probability given by
\begin{align}
P(z_m=1)&= C
e^{\langle\ln\pi_m\rangle}e^{-\frac{\gamma\langle(y_m-\boldsymbol{a}_m^r\boldsymbol{x})^2\rangle}{2}}
\label{z-update-1}\\
P(z_m=0)&=C e^{\langle\ln(1-\pi_m)\rangle} \label{z-update-2}
\end{align}
where $C$ is a normalizing constant such that
$P(z_m=1)+P(z_m=0)=1$, and
\begin{align}
\langle(y_m-\boldsymbol{a}_m^r\boldsymbol{x})^2\rangle
=&(y_m-\boldsymbol{a}_m^r\boldsymbol{\mu_x})^2+\boldsymbol{a}_m^r\boldsymbol{\Phi}_x{\boldsymbol{a}_m^r}^T
\nonumber\\
\langle\ln\pi_m\rangle=&\Psi(e+\langle z_m\rangle)-\Psi(e+f+1)
\nonumber\\
\langle\ln(1-\pi_m)\rangle=&\Psi(1+f-\langle
z_m\rangle)-\Psi(e+f+1)
\end{align}
The last two equalities can also be found in
\cite{PaisleyCarin09}, in which $\Psi(\cdot)$ represents the
digamma function.

\textbf{5) Update of $q_{\pi}(\boldsymbol{\pi})$:} The posterior
approximation of $q_{\pi}(\boldsymbol{\pi})$ can be calculated as
\begin{align}
\ln q_{\pi}(\boldsymbol{\pi})\propto&\langle\ln
p(\boldsymbol{z}|\boldsymbol{\pi})+\ln
p(\boldsymbol{\pi}|e,f)\rangle_{q_z(\boldsymbol{z})} \nonumber\\
\propto&
\sum_{m=1}^M\langle z_m\ln\pi_m+(1-z_m)\ln(1-\pi_m)+(e-1)\ln\pi_m \nonumber\\
&+(f-1)\ln(1-\pi_m)\rangle \nonumber\\
=& \sum_{m=1}^M\langle(z_m+e-1)\ln\pi_m+(f-z_m)\ln(1-\pi_m)\rangle
\end{align}
It can be easily verified that $q_{\pi}(\boldsymbol{\pi})$ follows
a Beta distribution, i.e.
\begin{align}
q_{\pi}(\boldsymbol{\pi})=\prod_{m}p(\pi_m)=\prod_{m}\text{Beta}(\langle
z_m\rangle+ e,1+f-\langle z_m\rangle) \label{pi-update}
\end{align}

In summary, the variational Bayesian inference involves updates of
the approximate posterior distributions for hidden variables
$\boldsymbol{x}$, $\boldsymbol{\alpha}$, $\boldsymbol{z}$,
$\boldsymbol{\pi}$, and $\gamma$ in an alternating fashion. Some
of the expectations and moments used during the update are
summarized as
\begin{align}
\langle\alpha_n\rangle=&\frac{\tilde{a}}{\tilde{b}_n}
\nonumber\\
\langle \gamma\rangle=&\frac{\tilde{c}}{\tilde{d}}
 \nonumber\\
\langle x_n^{2}\rangle=& \langle
x_n\rangle^2+\boldsymbol{\Phi}_x(n,n)
\nonumber\\
\langle z_m\rangle=&\frac{P(z_m=1)}{P(z_m=1)+P(z_m=0)} \nonumber
\end{align}
where $\boldsymbol{\Phi}_x(n,n)$ denotes the $n$th diagonal
element of $\boldsymbol{\Phi}_x$.

\section{Simulation Results}
We now carry out experiments to illustrate the performance of our
proposed method which is referred to as the beta-Bernoulli prior
model-based robust Bayesian compressed sensing method
(BP-RBCS)\footnote{Codes are available at
http://www.junfang-uestc.net/codes/RBCS.rar}. As discussed
earlier, another robust compressed sensing approach is
compensation-based and can be formulated as a conventional
compressed sensing problem (\ref{RCS-compensation}). For
comparison, the sparse Bayesian learning method
\cite{Tipping01,JiXue08} is employed to solve
(\ref{RCS-compensation}), and this method is referred to as the
compensation-based robust Bayesian compressed sensing method
(C-RBCS). Also, we consider an ``ideal'' method which assumes the
knowledge of the locations of the outliers. The outliers are then
removed and the sparse Bayeisan learning method is employed to
recover the sparse signal. This ideal method is referred to as
RBCS-ideal, and serves as a benchmark for the performance of the
BP-RBCS and C-RBCS. Note that both C-RBCS and RBCS-ideal use the
sparse Bayesian learning method for sparse signal recovery. The
parameters $\{a,b,c,d\}$ of the sparse Bayesian learning method
are set to $a=b=c=d=10^{-10}$. Our proposed method involves the
parameters $\{a,b,c,d,e,f\}$. The first four are also set to
$a=b=c=d=10^{-10}$. The beta-Bernoulli parameters $\{e,f\}$ are
set to $e=0.7$ and $f=1-e=0.3$ since we expect that the number of
outliers is usually small relative to the total number of
measurements. Our simulation results suggest that stable recovery
is ensured as long as $e$ is set to a value in the range
$[0.5,1]$.

We consider the problem of direction-of-arrival (DOA) estimation
where $K$ narrowband far-field sources impinge on a uniform linear
array of $M$ sensors from different directions. The received
signal can be expressed as
\begin{align}
\boldsymbol{y}=\boldsymbol{A}\boldsymbol{x}+\boldsymbol{w}
\nonumber
\end{align}
where $\boldsymbol{w}$ denotes i.i.d. Gaussian observation noise
with zero mean and variance $1/\gamma$,
$\boldsymbol{A}\in\mathbb{C}^{M\times N}$ is an overcomplete
dictionary constructed by evenly-spaced angular points
$\{\theta_n\}$, with the $(m,n)$th entry of $\boldsymbol{A}$ given
by $a_{m,n}=\exp \left\{-2j\pi(m-1)\sin
(\theta_n)D/\lambda\right\}$, in which $D$ denotes the distance
between two adjacent sensors, $\lambda$ represents the wavelength
of the source signal, and $\{\theta_n\}$ are evenly-spaced grid
points in the interval $[-\pi/2,\pi/2]$. The signal
$\boldsymbol{x}$ contains $K$ nonzero entries that are
independently drawn from a unit circle. Suppose that $T$ out of
$M$ measurements are corrupted by outliers. For those corrupted
measurements $\{y_m\}$, their values are chosen uniformly from
$[-10,10]$.


\begin{figure}[!t]
 \centering
\begin{tabular}{cc}
\hspace*{-3ex}
\includegraphics[width=4.9cm,height=4.9cm]{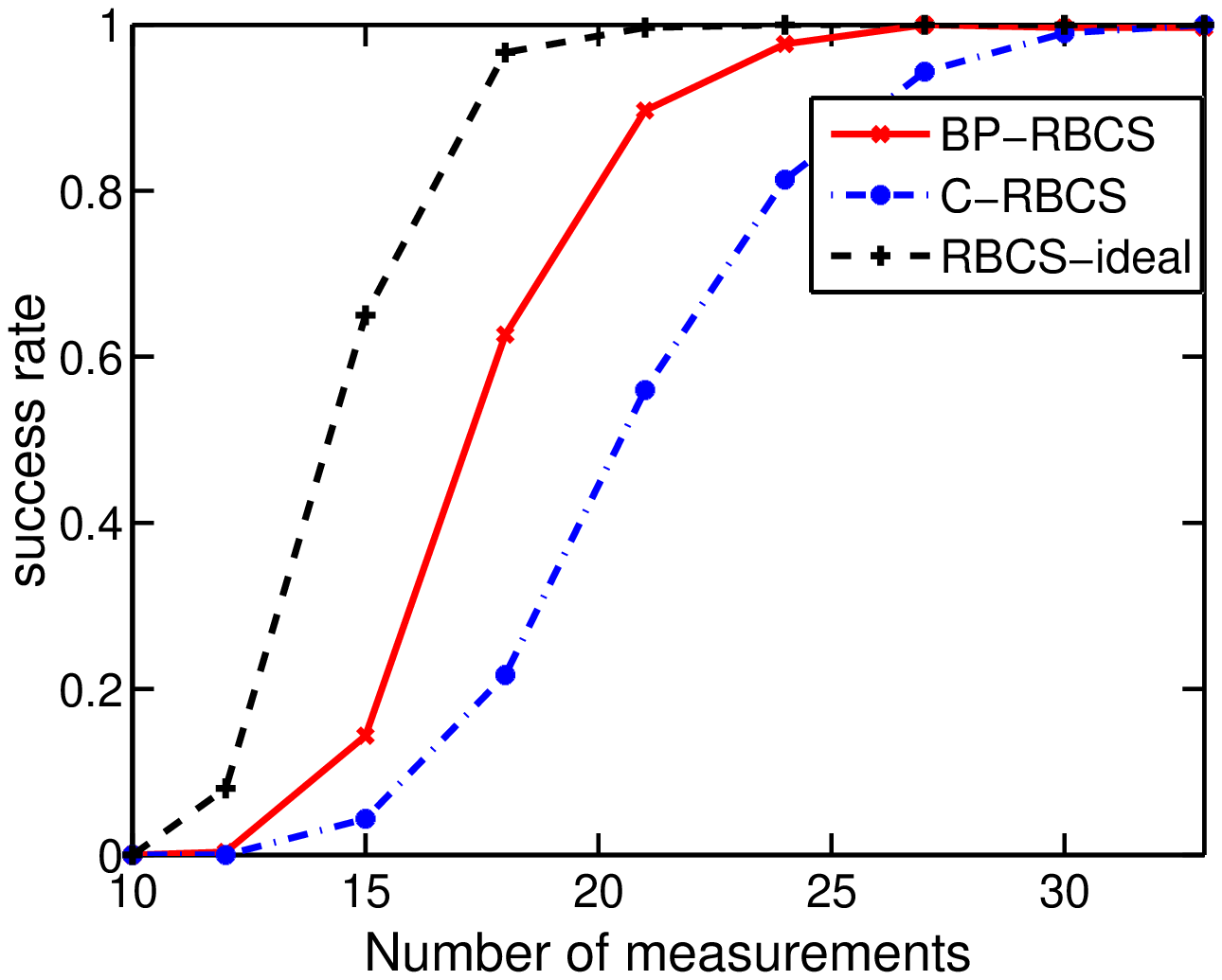}&
\hspace*{-5ex}
\includegraphics[width=4.9cm,height=4.9cm]{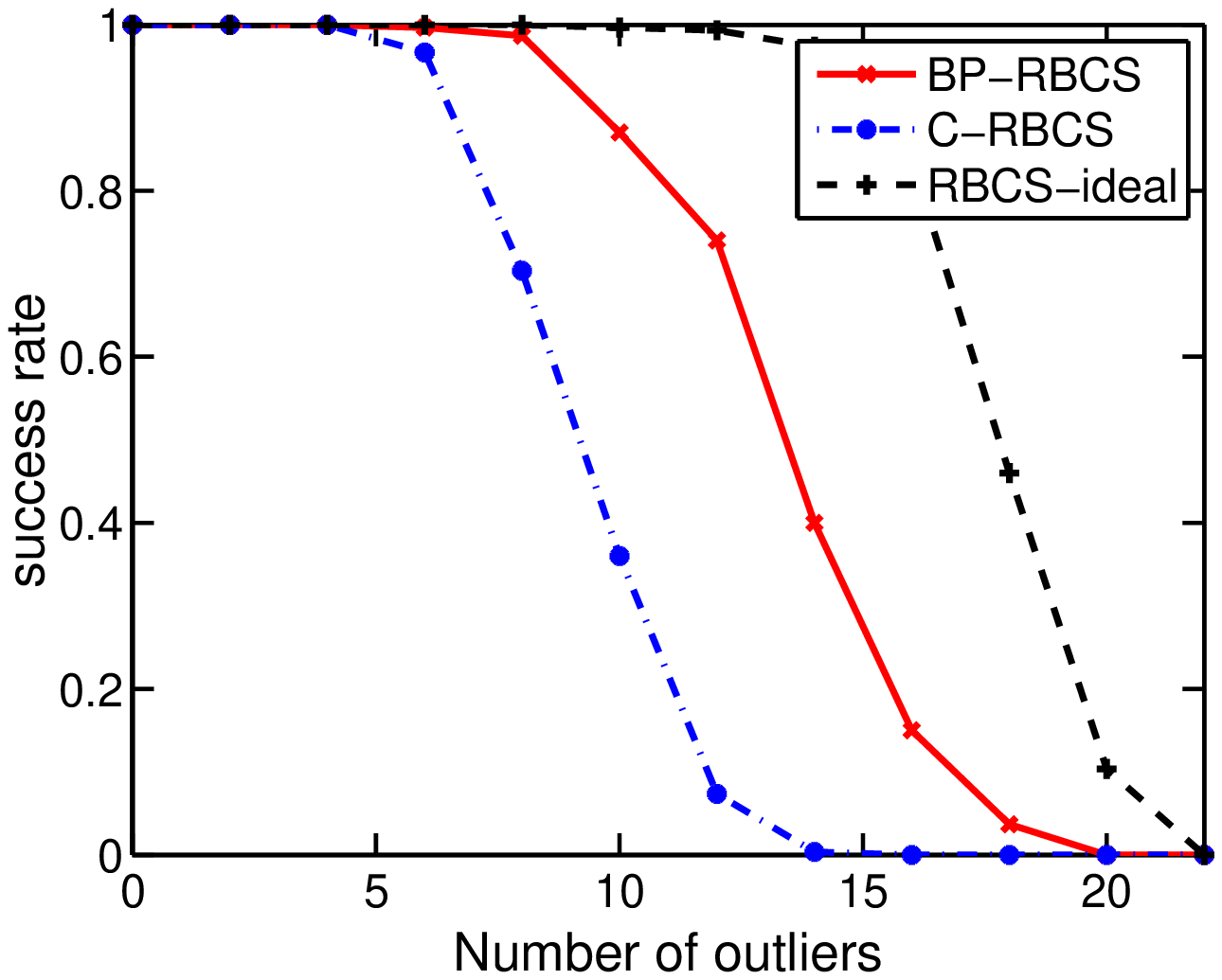}
\\
(a)& (b)
\end{tabular}
  \caption{(a) Success rates of respective algorithms vs. $M$; (b). Success rates of respective algorithms vs. $T$.}
   \label{fig1}
\end{figure}

\begin{figure}[!t]
 \centering
\begin{tabular}{cc}
\hspace*{-3ex}
\includegraphics[width=4.9cm,height=4.9cm]{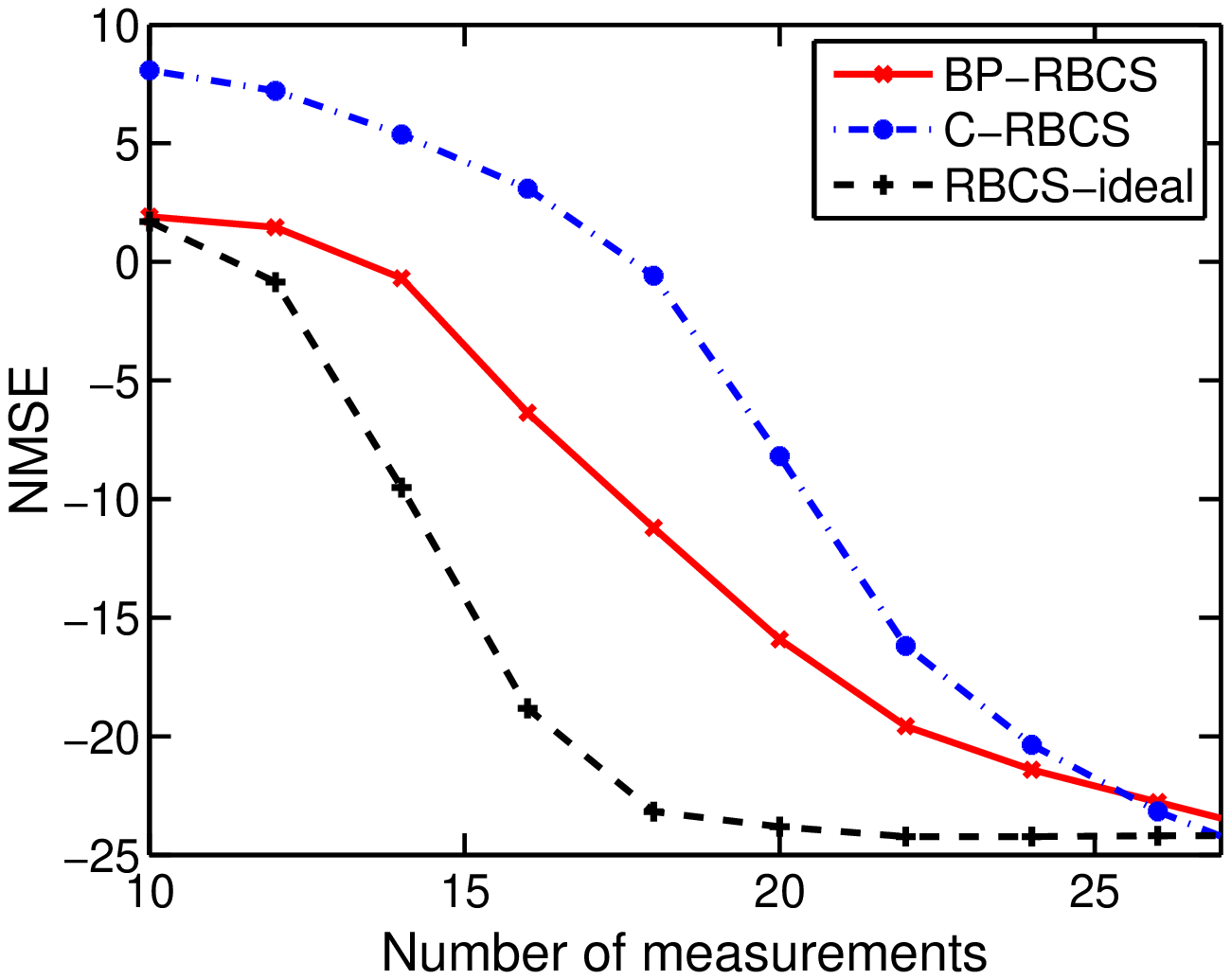}&
\hspace*{-5ex}
\includegraphics[width=4.9cm,height=4.9cm]{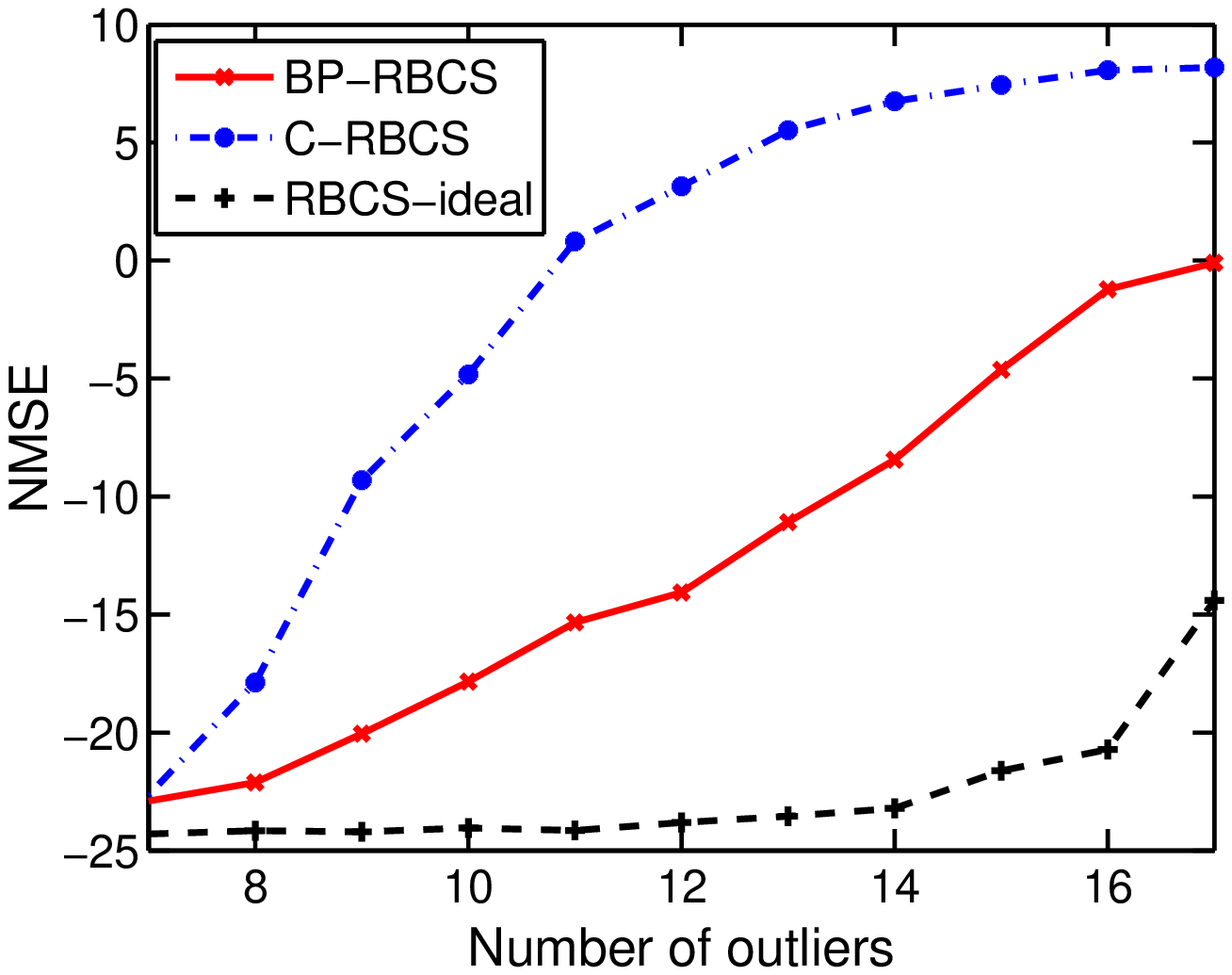}
\\
(a)& (b)
\end{tabular}
  \caption{(a) NMSEs of respective algorithms vs. $M$; (b). NMSEs of respective algorithms vs. $T$.}
   \label{fig2}
\end{figure}

We first consider a noiseless case, i.e. $1/\gamma=0$. Fig.
\ref{fig1} depicts the success rates of different methods vs. the
number of measurements and the number of outliers, respectively,
where we set $N=64$, $K=3$, $T=7$ (the number of outliers) in Fig.
\ref{fig1}(a), and $M=25$, $K=3$, $N=64$ in Fig. \ref{fig1}(b).
The success rate is computed as the ratio of the number of
successful trials to the total number of independent runs. A trial
is considered successful if the normalized reconstruction error of
the sparse signal $\boldsymbol{x}$ is no greater than $10^{-6}$.
From Fig. \ref{fig1}, we see that our proposed BP-RBCS achieves a
substantial performance improvement over the C-RBCS. This result
corroborates our claim that rejecting outliers is a better
strategy than compensating for outliers, particularly when the
number of measurements is small, because inaccurate estimation of
the compensation vector could lead to a destructive, instead of a
constructive, effect on sparse signal recovery. Next, we consider
a noisy case with $1/\gamma=0.01$. Fig. \ref{fig2} plots the
normalized mean square errors (NMSEs) of the recovered sparse
signal by different methods vs. the number of measurements and the
number of outliers, respectively, we set $N=64$, $K=3$, $T=7$ in
Fig. \ref{fig2}(a), and $M=25$, $K=3$, $N=64$ in Fig.
\ref{fig2}(b). This result, again, demonstrates the superiority of
our proposed method over the C-RBCS.



\section{Conclusions}
We proposed a new Bayesian method for robust compressed sensing.
The rationale behind the proposed method is to identify the
outliers and exclude them from sparse signal recovery. To this
objective, a set of indicator variables were employed to indicate
which observations are outliers. A beta-Bernoulli prior is
assigned to these indicator variables. A variational Bayesian
inference method was developed to find the approximate posterior
distributions of the latent variables. Simulation results show
that our proposed method achieves a substantial performance
improvement over the compensation-based robust compressed sensing
method.

\bibliography{newbib}
\bibliographystyle{IEEEtran}

\end{document}